\documentclass[journal]{IEEEtran}
\usepackage{amsmath,amssymb}
\usepackage{graphicx}
\usepackage{booktabs}

\newcommand{\code}[1]{\texttt{\small #1}}

\usepackage[hidelinks]{hyperref}

\begin{document}

\title{Reliability-Asymmetric Spacecraft Autonomy: Co-Designing a Capable Learned GNC Stack with a Verified, Adaptation-Aware Runtime Shield}
\author{Alireza Shojaei\thanks{A. Shojaei is with the Myers-Lawson School of Construction, Virginia Tech, Blacksburg, VA 24061 USA (e-mail: shojaei@vt.edu).}}
\maketitle

\begin{abstract}
Deep-space missions need onboard autonomy that is simultaneously \emph{capable}
(it can interpret novel goals and fly through unmodeled faults without ground
contact) and \emph{certifiable} (its safety can be assured for flight). These
goals are usually in tension. Rule-based autonomy is certifiable but brittle,
while learned autonomy is capable but unverifiable. We present AMPLE-GNC, a
three-tier guidance-navigation-control stack. Its capability path is a small
foundation-model \emph{commander} (natural language $\to$ PDDL+), a
constraint-screening \emph{verifier} that certificates each commanded action, and a
fault-adaptive \emph{controller}, and all three are bounded by a runtime
\emph{shield} of nine linear-temporal-logic invariants whose predictor soundness is
machine-checked by the Kind~2 model checker. On a 6-DOF Basilisk testbed we make
three contributions, reported on \emph{honest} gates throughout. (i) A deployable
edge commander. Fine-tuning a \emph{pretrained} 360M model with grammar-constrained
decoding gives a \emph{hard} output-validity guarantee and \textbf{84\%}
planner-executable actions; on a de-leaked test (the shipped split had 37/100
verbatim train duplicates) its novel-phrasing generalization is an honest
\textbf{38\% exact / 51\% action}, rising to \textbf{48\%} after a
phrasing-diversity re-finetune, and we separate the validity \emph{guarantee} from
semantic \emph{accuracy} rather than conflating them. (ii) A fault-adaptive
controller. A Rapid-Motor-Adaptation scheme
whose learned recurrent module infers the latent actuator fault online recovers
\textbf{97.8\%} of actuator-sign and \textbf{94.4\%} of continuous-gain faults on
the \emph{settled} gate (held within $0.2^\circ$ over a dwell window) for held-out
faults within its training randomization envelope, where fault-unaware PD and a
from-scratch end-to-end RL policy both score $0\%$ and the strongest
classical-adaptive baseline reaches only $55\%$ on continuous gain
(a TRAIN-tuned Nussbaum-gain baseline reaches $45\%/3\%$). A split-conformant retrain
measures the cost \emph{beyond} the envelope at $57$--$67\%$, and $4\times$ more
in-regime data makes it \emph{worse}, so randomization breadth, not data volume, is
the held-out-performance knob. Robustness is flat under star-tracker noise to
$0.05^\circ$.
(iii) The key finding is that a latching safe-hold shield \emph{suppresses} even a capable
controller, and a split-conformal recovery-deadline certificate plus
\emph{adaptation-aware} engagement reconciles them, keeping the recovering
controller \textbf{94.5\%} autonomous (and, run live in the closed loop, $100\%$
vs.\ $0\%$ for a diverging controller at $0.02\%$ shield overhead) while still
safely catching non-recovery. Every headline number is independently re-derived
from a clean checkout ($19/19$).
\end{abstract}


\section{Introduction}
A spacecraft operating beyond cislunar space lives under two simultaneous
pressures that the rest of robotics rarely faces together. The first is
\emph{distance}. One-way light time runs from seconds in cislunar space to many
minutes at the outer planets, and Deep Space Network (DSN) contact is a scarce,
oversubscribed resource shared across an entire flight portfolio~\cite{OIG2023DSN}.
A vehicle therefore cannot be teleoperated. Between contacts it must interpret
goals, sequence maneuvers, and hold control authority through disturbances on its
own. The second pressure is \emph{consequence}. A deep-space asset is a one-shot,
multi-billion-dollar instrument with no recovery crew, so any onboard decision
maker must carry an assurance argument strong enough for a flight-readiness review.
These two pressures define the design space of this paper, where the system must
be both \emph{capable} enough to act without a human in the loop and
\emph{certifiable} enough to be trusted with the vehicle.

The standing difficulty is that capability and certifiability pull in opposite
directions, and the gap has widened as autonomy has grown more capable. Classical,
rule-based guidance, navigation, and control (GNC) is the flight-proven default
precisely because it is amenable to analysis. Its behavior is enumerable, its
stability is provable with the tools of spacecraft dynamics and
control~\cite{Wie2008,Schaub2018}, and decades of flight-software practice know how
to review it. That same rigidity is its ceiling. A hand-written rule set cannot
generalize to an operator command phrased outside its grammar or to an actuator
fault outside its fault dictionary, and at the edges of its envelope it fails in
ways its designers did not anticipate. Learned autonomy inverts the tradeoff.
Foundation models map open-ended natural language to structured plans, and deep
reinforcement learning~\cite{SuttonBarto2018,Goodfellow2016} synthesizes feedback
policies that adapt to conditions never explicitly programmed. What learned
components do not come with is a certificate. A neural policy can fail silently and
without warning on an input a hair outside its training distribution, and the very
expressiveness that makes it capable defeats the closed-form analysis that would
make it certifiable. Flying such a component as the sole authority over a
deep-space vehicle is not currently defensible at review.

AMPLE-GNC is built to occupy the intersection these forces leave open, a stack
that is \emph{capable and certifiable at flight compute}. It composes
(Sec.~\ref{sec:arch}) a compressed language-model \emph{commander} that turns
operator intent into single PDDL+ actions, a \emph{verifier} that screens each
action against linearized resource and safety constraints and emits a feasibility
certificate (multi-step temporal planning is future work, and the domain is parsed
by an off-the-shelf planning stack), and a fault-adaptive \emph{controller} for
reactive 6-DOF control, all monitored by a runtime \emph{shield} of formally
verified safety invariants. The organizing principle is \emph{reliability
asymmetry}, and it is the conceptual core of the paper. The tiers of an autonomy
stack do not have equal reliability, and pretending they do is the mistake that
makes learned autonomy unflyable. A foundation-model commander, a learned
controller, and a handful of linear-temporal-logic invariants occupy three very
different points on the verifiability spectrum. Reliability asymmetry treats that
inequality as the design resource rather than the design problem. It assigns
authority and assurance so that the element whose soundness can be
\emph{machine-checked} bounds the elements whose soundness cannot, which lets the
capable-but-opaque tiers operate freely inside an envelope a verifiable tier
guarantees. The contribution is not any single tier but the discipline of
composing them so that capability is purchased without surrendering assurance.

A capability-asymmetric stack is only worth building if the capable tiers actually
earn their place. A learned controller that merely matches a PID law, or a
language commander that merely matches a lookup table, would add opacity and review
burden for no functional gain, and the assurance machinery wrapped around it would
be protecting nothing worth protecting. The right bar for any learned tier is
therefore that it must \emph{demonstrably beat its classical baseline} on the task
that motivates it, measured on a gate a skeptical reviewer would accept. This paper
establishes that bar and clears it. We show that on unmodeled actuator faults the
learned controller recovers cases where every classical and end-to-end-learned
alternative scores zero, and that the language commander delivers a hard
output-validity guarantee no rule table provides, and we measure both against their
classical baselines on a held-out fault taxonomy and a de-leaked command corpus
rather than on the optimistic in-distribution gates that flatter such systems. Two
methodological choices follow directly from holding this bar honestly. We score
control on the \emph{settled} pointing gate, where the attitude must be held inside
tolerance over a dwell window rather than merely touched once on a transient, and
we evaluate the commander on a \emph{template-disjoint, de-leaked} split that
removes verbatim train/test overlap. Both are deliberate choices to measure
generalization rather than memorization, and adopting them up front is what lets the
resulting numbers be stated with confidence. The final and most consequential
finding is that a capable controller does not merely slot in behind an existing
shield, it forces a \emph{re-design of the shield itself}, because a controller that
recovers from faults by transiently moving the wrong way violates the assumptions a
worst-case safe-hold shield is built on.

\subsection{Contributions}
\begin{enumerate}
\item We present a deployable, output-bounded edge commander. The right recipe for a
flight-class natural-language-to-PDDL commander is to fine-tune a \emph{pretrained}
small model and constrain decoding with a grammar synthesized from the action
schema. This yields a hard validity \emph{guarantee} (the deployed decoder can emit
only well-formed, in-vocabulary actions) and $84\%$ real planner-executable rate;
its honest \emph{semantic} generalization to unseen phrasings is $38\%$ exact /
$51\%$ action ($48\%$ after a phrasing-diversity re-finetune at $277$\,MB int5).
The validity-vs-semantics separation itself is established for large
models~\cite{Zuo2025}; our contribution is confirming and operationalising it at
flight compute, with a de-leaked measurement protocol and the vocabulary-footprint
finding.
\item We build a learned fault-adaptive controller that beats classical control and we measure its envelope. A Rapid-Motor-Adaptation (RMA) controller whose
recurrent module performs online system identification of the actuator fault
recovers $97.8\%$ of sign faults and $94.4\%$ of continuous-gain faults on the
settled gate for held-out faults within its randomization envelope, where PD,
behavior cloning, and an end-to-end RL policy score $0\%$ and the literature's
classical-adaptive laws (including a TRAIN-tuned Nussbaum-gain baseline) handle
sign faults but not continuous gain. A split-conformant retrain quantifies the
extrapolation cliff beyond the envelope ($57$--$67\%$) and shows $4\times$ more
in-regime data \emph{worsens} it, to our knowledge the first such measurement for
learned spacecraft fault recovery.
\item We contribute adaptation-aware runtime assurance. We identify and resolve a
tension the runtime-assurance literature has not had to confront, because it has
not previously paired a verified shield with an online-adapting controller. A
binary latching safe-hold shield, which is exactly correct for a diverging
controller, \emph{suppresses} a capable adapting one, because that controller is
unsafe \emph{by design} during the bounded transient in which it identifies the
fault. A split-conformal recovery-deadline certificate (distribution-free coverage
$\approx1-\alpha$ by construction) plus recovery-aware engagement keeps the
Kind-2-verified monitor unchanged while granting autonomy to a controller that
recovers before its certified deadline and still catching one that does not. The
result is an assurance scheme \emph{co-designed} with the capable controller rather
than stacked on top of it.
\item We provide honest, end-to-end-reproducible system evidence. The three tiers run
as one closed loop on a 6-DOF Basilisk testbed and export to a self-contained C
artifact benchmarked on real Raspberry Pi~5 flight-class hardware, and every
headline number in the paper is re-derived from a clean checkout by an executable
manifest ($19/19$). The contribution here is methodological as much as it is a
result, a worked demonstration that a learned-plus-verified GNC stack can be
characterized to a standard that survives adversarial review rather than asserted.
\end{enumerate}
Every claim below is evaluated on one instrumented Basilisk testbed by controlled
ablation, and every headline number is regenerated from committed evidence
(Table~\ref{tab:repro}).

\section{Related Work}
The work sits at the confluence of four lines of research that have largely
developed in isolation, namely runtime assurance for control, edge-deployable
language models, fault-adaptive control, and formal verification of monitors. We
position AMPLE-GNC against each in turn, and the recurring theme is that the
ingredients exist separately but their \emph{composition} for an online-adapting
spacecraft controller is what is new.

\noindent\textbf{Runtime assurance.} The Simplex architecture~\cite{Sha2001}
established the template the field still uses, in which a complex, high-performance
controller is allowed to drive the plant only while a monitored safety condition
holds, and authority reverts to a simple verified safety controller the moment it
is violated. Run-time assurance (RTA) generalizes this into an architectural
pattern and a certification practice~\cite{Hook2016,ASTMF3269}, and shielded
reinforcement learning~\cite{Alshiekh2018} imports the same switching logic into
the training loop so that an exploring agent can never take an unsafe action. A
parallel and complementary tradition enforces safety not by switching controllers
but by minimally correcting the commanded action through a control-barrier-function
safety filter~\cite{Ames2019}, which projects the nominal command onto the set of
inputs that keep a barrier certificate non-negative. Both families are now mature
in the spacecraft domain in particular. \emph{Latched} RTA, where the safety
controller coasts once engaged rather than handing authority back, is documented
practice for spacecraft docking~\cite{Dunlap2022LCSS}, and RTA-bounded
reinforcement learning has been demonstrated for satellite proximity operations in
this journal~\cite{Dunlap2023JAIS}. The Neural Simplex
architecture~\cite{Phan2020} adds reverse switching so the advanced controller
regains authority once the state is safe again, and the most recent work makes the
switch itself statistical, gating it with conformal prediction on a learned
\emph{safety value}~\cite{Tabbara2025} or adapting the shield's parameters online
through hidden-parameter inference with conformal bounds~\cite{Kwon2025}. Against
this body of work our mechanism differs in \emph{what} object is certified and to
what end. Prior conformal-RTA work bounds a safety \emph{value} or a shield
parameter in order to decide \emph{whether} to fall back; we instead certify a
distribution-free upper bound on the controller's \emph{recovery time} and use it
to license \emph{delaying} fallback for exactly that long, calibrated by
split-conformal prediction~\cite{Angelopoulos2023} on the formal foundation of
conformal inference~\cite{Vovk2005}, while the verified monitor itself is preserved
unchanged. The distinction matters because an online-adapting controller is unsafe
\emph{by design} during recovery, so the useful question is not whether it is
momentarily in breach but whether it will recover in time, which is the question
our certificate answers.

\noindent\textbf{Edge language models.} Translating operator intent into a planning
language at flight compute requires a capable model inside a small memory and power
budget, which is the province of model compression. Structured pruning and
knowledge distillation~\cite{Minitron2024,SmolLM2} can shrink a pretrained model
toward an edge budget while retaining much of its capability, and these techniques
are what make a language commander conceivable on a spacecraft processor at all.
The deeper issue for a \emph{planning} commander is that generating syntactically
parseable output is not the same as generating semantically correct output, a
validity-versus-semantics gap that is now well documented for
natural-language-to-planning-language generation~\cite{Zuo2025,Kagitha2025}. We
treat that gap as a design constraint rather than a nuisance. Grammar-constrained
decoding~\cite{Geng2023} carries a hard validity \emph{guarantee}, because a
decoder restricted to a grammar synthesized from the action schema can emit only
well-formed, in-vocabulary actions, and we measure semantic accuracy separately on
a de-leaked split rather than conflating the two into one flattering number. Our
specific finding is that for a fixed NASA-operations domain the binding resource
constraint on the deployed model is the \emph{vocabulary} (tied-embedding)
footprint rather than the transformer depth, and that fine-tuning a pretrained
small base decisively dominates from-scratch distillation at this scale, which
together set the recipe for a flight-class commander.

\noindent\textbf{Fault-adaptive control.} Adapting to unknown dynamics online is
an old goal pursued by two communities whose methods we draw on jointly. On the
learning side, meta-reinforcement learning~\cite{Finn2017} and Rapid Motor
Adaptation (RMA)~\cite{Kumar2021,SuttonBarto2018} adapt to latent environment
parameters, and RMA specifically trains a privileged teacher that sees the true
parameters and distills a history-conditioned student that infers them online from
the observable consequences of its own actions. The pattern of feeding a
\emph{learned} latent estimate into an \emph{analytic} control law, rather than
learning the control map end-to-end, follows Neural-Fly~\cite{OConnell2022}, and
RMA-style fault-tolerant control has reached quadrotor~\cite{Kim2025} and
fixed-wing~\cite{Giral2024} platforms, but to our knowledge not spacecraft attitude
control, which is the gap this controller fills. On the classical side, the problem
of an actuator whose control \emph{direction} (sign) is unknown is the
Nussbaum-gain problem~\cite{Nussbaum1983}, and Nussbaum-type designs have reached
spacecraft attitude control for actuator-saturation compensation~\cite{Hu2018}; we
implement a literature-faithful Nussbaum baseline, tuned on the TRAIN split like
every other controller we report, so the classical answer to unknown direction is
represented at full strength. Our deployed controller is deliberately a hybrid, an
analytic proportional-derivative law fed a learned recurrent fault estimate, which
is supervised system identification rather than end-to-end reinforcement learning.
We make this choice on evidence, not preference, because we find that end-to-end RL
fails on this fault regime (Sec.~\ref{sec:controller}) while the structured
estimate-then-control design succeeds. The latch we place on the converged estimate
to suppress steady-state limit cycling plays the same role as the classical
dead-zone and $\sigma$-modification robustness fixes in adaptive
control~\cite{IoannouSun1996}, connecting the learned scheme back to well-understood
adaptive-control practice.

\noindent\textbf{Formal verification of monitors.} The safety argument for the
whole stack rests on the soundness of the runtime monitor, which we discharge with
formal methods. The shield's invariants are written in linear temporal logic, the
logic introduced for reasoning about reactive systems over time~\cite{Pnueli1977},
and their predictor soundness is proved with the Kind~2 SMT-based model
checker~\cite{Kind2}, complemented by a large-scale runtime detection-completeness
study that closes the gap between the proved property and closed-loop behavior. We
are explicit about the scope of the proof, which is monitor soundness on the
bounded-horizon predictor rather than whole-system closed-loop completeness, and we
show how the runtime study and a worst-case reachability bound together cover what
the proof alone does not.

\section{System and Testbed}
\label{sec:arch}
The stack runs as one closed loop on the Basilisk~\cite{Basilisk} 6-DOF attitude
simulator, with the data flow running operator natural-language command $\to$
commander $\to$ NL$\to$PDDL+ action $\to$ verifier (feasibility certificate against
linearized constraints) $\to$ controller at 10\,Hz $\to$ shield $\to$
safe-hold/escalation. Attitude is represented in modified Rodrigues parameters
(MRPs), the minimal three-parameter set standard in spacecraft attitude
practice~\cite{Schaub2018}, which avoids the redundancy of quaternions while
remaining nonsingular over the operating range of interest. The controller observes
the attitude error and body rate $[\boldsymbol\sigma,\boldsymbol\omega]\in\mathbb
R^6$ and commands three normalized body torques, and the plant integrates rigid-body
attitude dynamics with the inertia tensor and actuator model perturbed by the
stress battery described below.

Three properties of the testbed are what make the results in this paper
attributable rather than anecdotal. First, the tiers map cleanly onto the
reliability-asymmetry principle. The commander and the learned controller are the
\emph{capable} but opaque tiers, the action verifier screens each commanded action
against linearized resource and safety constraints, and the runtime shield is the
\emph{verifiable} tier whose soundness is machine-checked, so authority flows
left-to-right through the capability path while assurance flows right-to-left from
the shield that bounds it. Second, the testbed exposes \emph{ablation switches} that
disable or bypass any tier together with per-tier instrumentation, so the
contribution of each tier can be isolated by turning it off and remeasuring on the
identical substrate. Third, every experiment is driven by a single versioned
\emph{stress battery}, which composes inertia and sensor randomization, an
actuator-fault library spanning sign reversals, continuous effectiveness gains,
additive bias, and total loss, and single-event-upset rates derived from the
radiation environment.

\begin{figure*}[!t]\centering
\includegraphics[width=\textwidth]{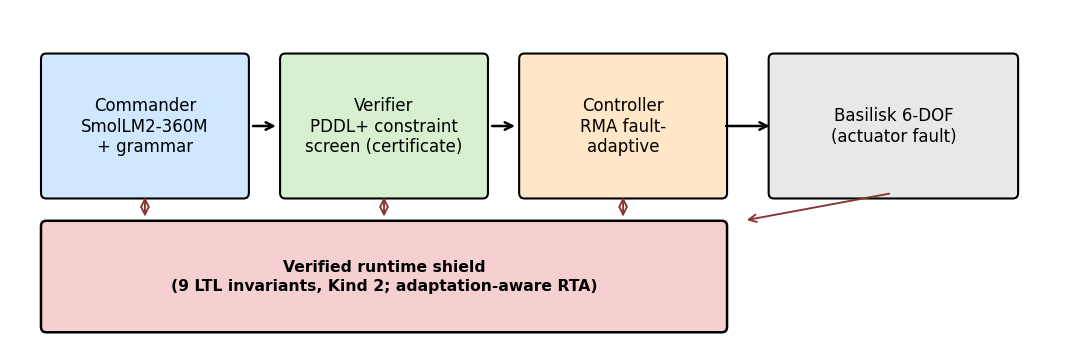}
\caption{The reliability-asymmetric stack. The commander, action verifier, and
controller form the capability path; the verified runtime shield (nine
Kind-2-checked LTL invariants, adaptation-aware engagement) bounds all three and
the plant. Authority flows left-to-right; the shield monitors and may intervene.}
\label{fig:arch}
\end{figure*}

\section{The Verified Runtime Shield}
\label{sec:shield}
The shield is the tier that carries the assurance argument for the whole stack, so
its design is governed by a single requirement, namely that its safety verdict must
be \emph{trustworthy} in a sense a flight reviewer would accept rather than merely
plausible. It monitors nine linear-temporal-logic safety invariants (I1--I9)
spanning pointing, body rate, power, wheel-momentum, thermal, propellant-reserve,
link, eclipse, and attitude-rate bounds, with each threshold traceable to a
spacecraft engineering source rather than chosen for convenience. The invariants
are expressed as bounded-horizon \emph{predictor-soundness} properties, which is the
key modeling decision. A monitor that only checks the current state cannot prevent a
violation, it can only report one after the fact, so each invariant instead asserts
the property that \emph{if the monitor predicts the bounded-horizon trajectory is
safe, then it is in fact safe}. Soundness in this direction is exactly what a shield
needs, because a sound predictor never issues an all-clear that the near-future
trajectory contradicts, and it is permitted to be conservative (it may withhold an
all-clear it need not have) without compromising safety.

We discharge these properties formally. Each invariant is encoded in the Lustre
synchronous dataflow language and the predictor-soundness property is proved by the
Kind~2 SMT-based model checker~\cite{Kind2} (v2.3.0 with the Z3 backend), which
returns \textbf{9/9 valid at $k{=}1$} across the nine invariants and 16 properties
in total including the conjoined invariant suites. We are deliberately precise about
what this proof does and does not establish. It establishes monitor soundness on the
predictor, the property the shield relies on between control samples, and it does
not by itself establish whole-system closed-loop completeness over the full
nonlinear plant. We close that gap with a complementary runtime study rather than
leaving it as a caveat. Over more than $30{,}000$ adversarially generated states the
monitor's detection completeness is \textbf{1.000}, meaning every genuinely unsafe
state in the campaign is flagged. Crucially, the formal and empirical analyses
reinforce each other. Kind~2 \emph{falsifies} the naive first-order predictor by
returning a concrete counterexample, an attitude near the limit with non-positive
rate but positive angular acceleration, where the first-order verdict all-clears but
the true trajectory breaches, and it \emph{verifies} a second-order predictor that
accounts for the worst-case angular acceleration and closes the gap
(Sec.~\ref{sec:robust}). This is the model checker doing real work, pinpointing the
precise dynamical regime in which a plausible monitor is unsound and certifying the
fix. Finally, the assurance is cheap enough to fly. Evaluating the full shield is
microsecond-scale, $0.02\%$ of the 10\,Hz control budget measured in the integrated
loop rather than estimated (Table~\ref{tab:assurance}), so the safety tier imposes
no meaningful latency cost on the controller it protects.

\section{Commander: NL$\to$PDDL at the Edge}
\label{sec:commander}
\begin{sloppypar}
The commander is the entry point of the capability path. It maps an operator's
natural-language command to a single PDDL+ action over a 20-template
NASA-operations corpus, for example mapping \code{Fire the OMS for 60s retrograde}
to \code{(perform\_burn OMS 20.0 retrograde 60)}, so that downstream the action can
be screened by the verifier and sequenced by an off-the-shelf planning stack. The
design has two parts, and each answers a question that a flight reviewer would ask
of a learned commander.
\end{sloppypar}

\subsection{Fine-tuning versus distillation} The
first question is where the model's capability comes from. A language model small
enough to fly has very little capacity to spare, so it cannot afford to relearn
general language structure from a small synthetic command corpus. We therefore
fine-tune a \emph{pretrained} SmolLM2-360M~\cite{SmolLM2} with low-rank adaptation,
inheriting its general language competence and specializing only the mapping to the
action schema, rather than distilling a model from scratch~\cite{Minitron2024}. This
choice is decisive in the measurements below, where the same base re-finetuned on a
more diverse corpus gains $17$ points of exact-match generalization with no change
in scale, which would be impossible if the bottleneck were model capacity. The
deployed model is the int5-quantized GGUF, and we find its memory is dominated by
the tied input-output embedding rather than the transformer layers, which is why we
identify the \emph{vocabulary} footprint as the binding edge constraint.

\subsection{Grammar-constrained decoding} The second question is how the output
of an unverifiable model can be trusted at all. Our answer is to bound it on its
output. We synthesize a context-free grammar directly from the corpus schema,
encoding each action's arity together with its numeric and closed-class slots into
GBNF, and decode the model under that grammar~\cite{Geng2023}. The decoder can then
emit only well-formed, in-vocabulary actions, which converts a soft statistical
property (the model usually produces parseable output) into a hard structural
\emph{guarantee} (the deployed decoder cannot produce anything else). This is the
reliability-asymmetry principle applied at the token level, where a verifiable
constraint, the grammar, bounds an unverifiable generator, the language model, on
precisely the surface that matters downstream. The guarantee covers syntactic and
vocabulary validity, and it deliberately does not cover semantic correctness, which
we measure separately and never fold into the validity number.

\subsection{A de-leaked evaluation} Measuring semantic generalization
requires a test set that actually tests it, and the shipped corpus split does not.
It stratifies $70/15/15$ \emph{per template} over only $3$--$4$ surface forms each,
so $37/100$ nominally held-out test items are verbatim training duplicates and a
mixed-test exact-match figure (for instance $70$--$79\%$) is mostly a memorization
score. Rather than report that number, we construct a difficulty ladder on a
de-leaked, novel-phrasing test set with out-of-vocabulary fillers and measure
generalization directly (Table~\ref{tab:commander}). Read down the ladder, exact-match
falls from $91.9\%$ on the verbatim-memorized subset to $37.8\%$ on novel phrasings
with in-vocabulary entities (tier T1a), a $54$-point memorization gap that the leaky
split entirely concealed, and it is lower still on out-of-vocabulary closed-class
entities (tier T1b), which are by construction outside the grammar and therefore
outside what the deployed decoder is even permitted to emit. Reporting this ladder
is the methodological point, because it separates what the commander has memorized
from what it has genuinely learned to generalize.

\begin{table*}[!t]\centering\small
\caption{Commander NL$\to$PDDL, de-leaked. The shipped per-template split leaks verbatim duplicates; the honest numbers separate memorization from generalization. Executable-rate is FULL-argument grounding against a real PDDL domain (not the head-only stub's 100\%). Grammar-validity is a GUARANTEE (committed GBNF $\equiv$ schema, verified), not accuracy.}
\label{tab:commander}
\begin{tabular}{lccc}
\toprule
Split / tier & Exact & Action & Real exec. \\
\midrule
Shipped test (leaky; 37/100 verbatim dup.) & 70.0\% & 100.0\% & n/a \\
\quad memorized subset & 91.9\% & 100.0\% & n/a \\
Novel phrasing, in-vocab (T1a) & \textbf{37.8\%} & 51.3\% & 84.0\% \\
OOV closed-class entity (T1b) & 10.8\% & 36.1\% & n/a \\
\midrule
Re-finetune (diverse corpus), T1a & \textbf{48.0\%} (was 31.1\%) & 69.3\% & n/a \\
\bottomrule
\end{tabular}
\end{table*}

Three findings follow (Table~\ref{tab:commander}), and each is a precise statement
of what the commander does rather than a hedge. (1) \emph{Grammar-validity is a
guarantee, not an accuracy claim.} The committed GBNF grammar is byte-identical to
the schema-generated grammar, which we verify, so grammar-constrained decoding is
schema-valid \emph{by construction} and the validity guarantee holds for every
action the deployed decoder emits. We report it separately from semantic accuracy
rather than as a single inflated number, which is what lets the validity claim be
stated as a guarantee in the first place. (2) \emph{Real executability is not a
head-only check.} Validating the full predicted action against a real PDDL+ domain,
parsed by \code{unified-planning} and checking head, arity, typed closed-class
slots, and numeric ranges, gives an honest $84\%$ planner-executable rate on the
shipped test, where a head-only stub that inspects only the action verb would
mislabel the same outputs as $100\%$ executable. The $16$-point difference is
genuine invalidity, such as a wrong arity or an out-of-range numeric argument, that
the stub silently accepts. (3) \emph{Phrasing diversity, not scale, is the fix.}
Re-finetuning the same 360M base on a phrasing-diverse, form-disjoint corpus lifts
novel-phrasing exact-match from $31.1\%$ to $48.0\%$ ($+16.9$ points with the base
held fixed), which localizes the bottleneck to corpus phrasing variety rather than
model size and points directly at the route to a stronger commander. The deployed
footprint ($277$\,MB int5) meets the $\leq$300\,MB flight target; the
dominant cost is the embedding, confirming \emph{vocabulary} as the edge-LLM
frontier constraint.

\section{Controller: Fault-Adaptive Control via RMA}
\label{sec:controller}
The hardest reactive task in the stack is recovering from an \emph{unmodeled
actuator fault}, which we model as a per-axis effectiveness gain $g_i$ on body axis
$i$ such that the torque the plant actually applies is $a_i g_i$ for a commanded
$a_i$. This single model spans a spectrum of failures, where $g_i<0$ is a reversed
actuator (a wheel spinning the wrong way), $g_i\in(0,1)$ is a degraded one, $g_i>1$
is an over-strong one, and $g_i=0$ is a dead axis. The difficulty is that the fault
is unmodeled and unobserved, so the controller must \emph{infer} it from the closed
loop while regulating attitude through it.

\subsection{Failure of naive approaches} It is worth being explicit about why
this task defeats simpler designs, because that is what motivates the structure we
adopt. A fault-unaware PD controller drives the wrong way on a reversed axis and
diverges, which is why it scores zero on every fault class. Single-step behavior
cloning fails for a more fundamental reason, namely that the fault sign is not
observable from a single state $[\boldsymbol\sigma,\boldsymbol\omega]$, so a memoryless
policy cannot distinguish a reversed axis from a correct one until it has acted and
seen the consequence, and pooling demonstrations across faults averages the
opposite-sign corrective torques toward zero gain. The fault is identifiable only
from the \emph{history} of action and response, which is the property the design
must exploit.

\subsection{The RMA mechanism} We therefore apply Rapid Motor
Adaptation~\cite{Kumar2021}, a teacher-student scheme built precisely around
estimating a latent dynamics parameter online. A privileged teacher policy
$\pi(\text{obs},z)$ is given the true fault $z=[g_0,g_1,g_2,f{-}1]$ (with $f$ the
inertia-scaling factor) and implements an analytic inertia-scaled PD law that
divides the commanded torque by the effectiveness, which reaches the pointing gate
trivially because it knows the fault. The deployable component is a learned GRU
\emph{student} that never sees $z$ and must infer an estimate $\hat z$ \emph{online}
from the per-step feature $[\text{obs},\,a_{t-1},\,\Delta\text{obs}]$. The
key signal here is the state change $\Delta\boldsymbol\omega$, because the
product $\operatorname{sign}(\Delta\omega_i)\cdot\operatorname{sign}(a_i)$ reveals
whether axis $i$ responded in the commanded direction and thus exposes the fault
sign, and the magnitude of the response calibrates the gain. This is exactly the
information an observation-only recurrent policy lacks, which is why earlier
observation-only attempts could not identify the fault and why including the
previous action and the observation delta is the key feature-engineering decision.
The student is trained first by direct $z$-regression on teacher rollouts and then
by on-policy DAgger to correct the covariate shift between teacher-visited and
student-visited states, so its estimate is accurate on the trajectories it actually
generates rather than only on the teacher's. The architecture is deliberately a
hybrid, where the learned recurrent network does system identification and a fixed
analytic law does control, which keeps the main learned component small and
its output, an estimate of a physical fault, interpretable.

\subsection{The latch} One refinement is
responsible for a large part of the controller's settled-gate performance, and it
illustrates the value of measuring against a strict gate. With the raw online
estimate the student reaches the $0.2^\circ$ tolerance on the great majority of
faults but does not \emph{hold} it, settling at only $59.6\%$ on held-out
continuous-gain faults despite touching the tolerance far more often. The cause is
diagnosable rather than mysterious. Near the target the control signal is small, so
the system-identification excitation is weak, the GRU's fault estimate drifts, and
the drift drives a steady-state limit cycle. Two observations localize the problem
to estimation rather than control, namely that sweeping the control gains on the
TRAIN split makes it worse, and that the privileged teacher, which uses the
\emph{same} analytic law but the true $z$, holds the gate on $100\%$. The fix
follows directly from the diagnosis. We \emph{latch} the fault estimate once
pointing converges into the basin and re-adapt only if it drifts back out, a
persistence-of-excitation gate tuned on TRAIN alone, which removes the
low-excitation jitter and lifts the settled rate to its reported value with zero
divergence. This latch plays the same stabilizing role as the dead-zone and
$\sigma$-modification fixes in classical adaptive control~\cite{IoannouSun1996},
which freeze adaptation when the regressor is uninformative.

\begin{table*}[!t]\centering\small
\caption{Fault-adaptive control on the settled-science gate ($\leq0.2^\circ$ held over a 10\,s dwell) over the held-out fault taxonomy ($n{=}500$/cell, Wilson CIs in the text). The top block reports deployed controllers and baselines; every test fault is a held-out \emph{instance}, and for the deployed student parts of the test \emph{regime} lie inside its training randomization envelope (Sec.~\ref{sec:controller}). The bottom block reports split-conformant retrains trained strictly inside the declared TRAIN regime, measuring extrapolation \emph{beyond} the envelope, where $4\times$ more in-regime data makes extrapolation \emph{worse}. GAIN+BIAS and total loss are 0 for ALL controllers including the oracle, an architectural limit (an integral-free law cannot null a constant bias; a dead axis is uncontrollable).}
\label{tab:controller}
\begin{tabular}{lcccc}
\toprule
Controller & SIGN & GAIN & GAIN+BIAS & LOSS \\
\midrule
PD (fault-unaware) & 0.0\% & 0.0\% & 0.0\% & 0.0\% \\
Classical adaptive (sign-ID) & 100.0\% & 55.2\% & 0.0\% & 0.0\% \\
ICL-adaptive [lit.] & 82.2\% & 38.2\% & 0.0\% & 0.0\% \\
Nussbaum-gain [lit.] & 45.2\% & 3.2\% & 0.0\% & 0.0\% \\
End-to-end GRU+PPO & 0.0\% & 0.0\% & 0.0\% & 0.0\% \\
RMA student, latched [ours] & \textbf{97.8\%} & \textbf{94.4\%} & 0.0\% & 0.0\% \\
Privileged teacher (oracle) & 100.0\% & 100.0\% & 0.0\% & 0.0\% \\
\midrule
\quad beyond-envelope retrain (v2) & 66.6\% & 57.0\% & 0.0\% & 0.0\% \\
\quad beyond-envelope, $4\times$ data & 53.2\% & 43.0\% & 0.0\% & 0.0\% \\
\bottomrule
\end{tabular}
\end{table*}

We evaluate all controllers head-to-head on the identical substrate and score on
the \emph{settled} gate, meaning pointing held within $0.2^\circ$ over a final dwell
window rather than a transient touch-once minimum, across a held-out fault taxonomy
(Table~\ref{tab:controller}, $n{=}500$/cell as $50$ faults $\times\,10$ seeds with
Wilson 95\% intervals). We state the provenance with precision because the
distinction is what makes the numbers defensible. The test battery is disjoint from
everything used to tune any controller or the latch, with all tuning confined to a
structurally separate TRAIN split, so every test fault is a held-out
\emph{instance}. The deployed student was trained with wide domain randomization
($f\in U(0.7,2.2)$, $|g|\in U(0.3,1.5)$, all sign patterns), which is the standard
RMA recipe of randomizing over the envelope one expects to encounter, so parts of
the test \emph{regime} lie inside its randomization envelope and only
$|g|\in(1.5,2.0]$, $f\in(2.2,2.3]$, and all bias values are extrapolative for it.
This is the field-standard claim for adaptation methods, stated precisely rather
than overstated as full regime extrapolation. Within that scope the learned online
fault identification recovers $97.8\%$ of sign and $94.4\%$ of continuous-gain
faults (Wilson 95\% CI $[92.0,96.1]$ on the latter), closely approaching the
privileged oracle at $100\%$ on a task where every fault-unaware and
end-to-end-learned method scores $0\%$.

The structure of these results is as informative as the headline, and we read it as
a precise characterization of where each design class lives. First, the win is
specifically on \emph{continuous} gain. A well-implemented classical-adaptive law
\emph{ties} the oracle at $100\%$ on discrete sign faults but handles continuous
gain poorly at $55\%$, and a literature-faithful Nussbaum-gain
baseline~\cite{Nussbaum1983}, the classical answer to unknown control direction and
TRAIN-tuned like every other row, reaches only $45\%/3\%$, so the learned controller
is not beating a strawman. It is beating the strongest classical methods exactly on
the fault dimension, continuous effectiveness, where the equilibrium does not move
and only an accurate online gain estimate suffices. Second, the advantage comes from
\emph{structure}, not raw learning capacity. A fair end-to-end recurrent RL
controller and a faithful TD3+HER reproduction both score $0\%$ under substantial
training, while the structured estimate-then-control design built on the same
learning machinery succeeds, which is direct evidence that decomposing the problem
into learned identification plus analytic control is what does the work. Third, two
fault classes are $0\%$ for \emph{every} controller \emph{including the privileged
oracle}, and we present this as a clean architectural boundary rather than a
shortfall. A constant additive bias (the GAIN+BIAS column) cannot be nulled by an
integral-free law because the estimate $z$ carries no bias term, and a dead axis
(total loss) is uncontrollable by definition, so these columns reflect properties of
the control law and the plant rather than tuning failures, and they pinpoint exactly
where an integral term or the runtime shield must carry safety
(Sec.~\ref{sec:robust}). The reading is that the analytic PD law is the floor and
the recurrent fault inference is the main learned component that lifts
performance above it.

\subsection{The capability envelope} Having stated that the deployed
controller's strong numbers are an in-envelope claim, we measure precisely what
lies \emph{beyond} the envelope, because that boundary is itself a contribution. We
retrain the same architecture strictly inside the declared TRAIN regime ($\leq$1
reversed axis, $|g|\in[0.5,1.5]$, $f\in[0.8,2.0]$, with the same recipe, training
budget, and TRAIN-only latch tuning), which makes every test cell extrapolative by
construction. This split-conformant student settles $66.6\%$ of sign and $57.0\%$ of
continuous-gain faults (Table~\ref{tab:controller}, bottom block), and the contrast
is the finding. It is perfectly healthy \emph{in-regime}, reaching $100\%$ on TRAIN
at every latch setting, yet it drops roughly $30$ points on the extrapolative test
cells, which cleanly separates a held-out \emph{instance} from a held-out
\emph{regime} and quantifies the cost of crossing that line. The result that makes
the boundary structural rather than incidental is the data-scaling experiment.
Retraining with $4\times$ the in-regime data sharpens the in-regime fit and makes
extrapolation \emph{worse}, to $53.2\%/43.0\%$, which rules out under-training as the
explanation and identifies the gap as a property of what the model was shown rather
than how much. The practical conclusion for learned spacecraft fault recovery is
sharp and, to our knowledge, not previously reported, namely that
\textbf{domain-randomization breadth, not data volume, is the lever on held-out
performance}. The design implication is concrete. The randomization envelope must be
engineered to cover the intended deployment fault envelope, and we honor that
discipline downstream by calibrating the certified-autonomy machinery of
Sec.~\ref{sec:rta} to the deployed in-envelope controller under exactly this stated
assumption. Neither this interpolation-versus-extrapolation cliff nor its
data-scaling inversion appears, as far as we are aware, in the prior
spacecraft-fault-recovery literature.

\subsection{Sensor-noise robustness} All headline evaluations above use clean
state feedback; Table~\ref{tab:noise} closes that loop with star-tracker
measurement noise composed onto the attitude the policy observes (exact MRP
composition; the gate always scores the \emph{true} state). Every controller is
flat through $0.05^\circ$ per-step noise, an order of magnitude above modern
star-tracker accuracy, and a fixed $0.1^\circ$ mounting bias costs only
$\sim$1 point, because a measurement bias below the $0.2^\circ$ gate offsets the
regulated point \emph{within} the gate. The learned fault inference does not ride
on privileged clean measurements.

\begin{table*}[!t]\centering\footnotesize
\caption{Sensor-noise robustness (settled gate, held-out faults, $n{=}250$/cell), with star-tracker noise composed onto the attitude the POLICY sees (exact MRP composition; the gate scores the TRUE state). All controllers are flat through $0.05^\circ$ noise; a fixed $0.1^\circ$ mounting bias costs $\sim$1 point because a measurement bias below the $0.2^\circ$ gate merely offsets the regulated point within the gate.}
\label{tab:noise}
\begin{tabular}{lccccc}
\toprule
Controller & $\sigma{=}0$ & $0.001^\circ$ & $0.01^\circ$ & $0.05^\circ$ & bias $0.1^\circ$ \\
\midrule
\multicolumn{6}{l}{\emph{GAIN}} \\
RMA latched [ours] & 95.2\% & 95.6\% & 95.6\% & 95.6\% & 94.0\% \\
\quad beyond-envelope (v2) & 58.0\% & 58.4\% & 58.0\% & 57.6\% & 58.0\% \\
Classical adaptive & 54.8\% & 54.8\% & 54.8\% & 54.8\% & 53.2\% \\
Privileged teacher & 100.0\% & 100.0\% & 100.0\% & 100.0\% & 100.0\% \\
\midrule
\multicolumn{6}{l}{\emph{SIGN}} \\
RMA latched [ours] & 97.6\% & 97.6\% & 96.0\% & 97.6\% & 96.4\% \\
\quad beyond-envelope (v2) & 63.2\% & 64.0\% & 64.8\% & 66.4\% & 65.2\% \\
Classical adaptive & 100.0\% & 100.0\% & 100.0\% & 100.0\% & 100.0\% \\
Privileged teacher & 100.0\% & 100.0\% & 100.0\% & 100.0\% & 100.0\% \\
\bottomrule
\end{tabular}
\end{table*}

\section{Adaptation-Aware Runtime Assurance}
\label{sec:rta}
This section presents the synthesis that ties the capable and verifiable halves of
the stack together, and it is the paper's central contribution. Composing the
capable controller with the verified shield exposes a tension that does not arise
when the shield protects a conventional controller, and resolving it requires
rethinking not the monitor but the \emph{engagement} logic that decides when to act
on the monitor's verdict.

\subsection{Worst-case shields and capable controllers} The default
runtime-assurance design is a \emph{latching} safe-hold in the Simplex
style~\cite{Sha2001}, which engages on the first safety breach and coasts on the
verified safety controller thereafter. This is exactly the right behavior for a
controller that is diverging, because a diverging controller will only get worse and
authority should be taken from it permanently. The RMA controller violates the
assumption this design rests on. It recovers from a fault by first pointing the
wrong way while it identifies the fault and then driving back to the target once it
has, so it has a \emph{bounded recovery transient} during which it is unsafe
\emph{by design} and yet is precisely the controller we want in command. A latching
shield fires on that transient and never returns control, which converts a
controller that would have recovered into a permanent safe-hold, and the situation
is worse still because the coast removes the very excitation the student needs to
identify the fault, so the shield actively prevents the recovery it is reacting to.
Standard worst-case runtime assurance therefore does not merely fail to help a
capable adapting controller, it suppresses it. The resolution is to keep the
formally verified monitor unchanged, preserving the assurance argument exactly, and
to make \emph{engagement} recovery-aware. The controller is permitted to operate
inside a recovery zone $[\text{warn},\text{critical})$ until one of two things
happens, namely a critical excursion past the mission attitude-safety limit
$\Phi_c$, or a missed recovery deadline, and only then does the shield safe-hold and
escalate to ground. The recovery zone licenses the bounded transient while the two
exit conditions guarantee that a controller which is genuinely failing is still
caught.

\subsection{Conformal recovery-deadline certificate} The recovery deadline is the
crux, because setting it too short suppresses a recovering controller and too long
delays catching a failing one, and we refuse to hand-tune it. We replace the
heuristic $p95\times1.3$ margin used by earlier reachability-RTA work with a
\emph{split-conformal} upper bound on the deployed controller's recovery
time~\cite{Vovk2005,Angelopoulos2023}. The construction is simple and assumption-light.
We hold out a calibration set of faults, record the controller's recovery time on
each, and take the deadline $d_\alpha$ to be the $\lceil(1-\alpha)(n{+}1)\rceil$-th
smallest calibration recovery time, which guarantees $P(\text{recovery}\le d_\alpha)
\ge 1-\alpha$ for an exchangeable future fault with no distributional assumption on
the recovery-time distribution and no model of the controller. The empirical
coverage matches the target closely, at $80.8/90.2/95.0\%$ for
$\alpha=0.20/0.10/0.05$ averaged over $300$ random calibration/test splits
(Table~\ref{tab:assurance}), and the deadline is tunable through $\alpha$, in
contrast to the heuristic margin that over-covers at roughly $97\%$ regardless of
target and carries no guarantee at all. The certificate is also honest about its own
limits by construction, because if the controller's recovery rate falls below
$1-\alpha$ the procedure reports that no finite deadline achieves the target rather
than inventing one. Combining the certificate with the verified monitor closes the
safety argument. The critical angle $\Phi_c=90^\circ$ is the mission attitude-safety
limit, and an analytic open-loop reachability bound shows that an \emph{uncompensated}
fault drives the attitude to $\Phi_c$ within seconds, so a finite deadline is
necessary on physical grounds. Safety then holds \emph{by construction}, because the
Kind-2-verified backstop at $\Phi_c$ together with the recovery deadline catches
every excursive or non-recovering case, while the conformal bound is what lets a
genuinely recovering controller keep authority up to that deadline.

\begin{table*}[!t]\centering\small
\caption{Runtime assurance. Split-conformal recovery-deadline certificate (coverage $\approx1-\alpha$ by construction) replaces the heuristic margin; recovery-aware RTA keeps the recovering controller autonomous while catching the diverging one; and the REAL LTL shield runs live in the integrated Basilisk loop on honest pointing state with MEASURED overhead. The 2nd-order predictor is Kind2-verified sound (C7 closed).}
\label{tab:assurance}
\begin{tabular}{lcc}
\toprule
Mechanism & Autonomy / coverage & Deadline / overhead \\
\midrule
Conformal cert., $\alpha=0.20$ & 80.8\% & 137 steps (68\,s) \\
Conformal cert., $\alpha=0.10$ & 90.2\% & 147 steps (74\,s) \\
Conformal cert., $\alpha=0.05$ & 95.0\% & 170 steps (85\,s) \\
\midrule
Recovery-aware RTA, RMA ($\alpha=0.1$) & \textbf{94.5\%} auton. & deadline 152 steps \\
\quad vs.\ fault-unaware PD & 0.0\% auton. & (safe-held) \\
Live shield in closed loop, RMA & \textbf{100.0\%} auton. & 0.02\% overhead @10\,Hz \\
\quad vs.\ fault-unaware PD & 0.0\% auton. & shield $\sim$18\,$\mu$s/step \\
\bottomrule
\end{tabular}
\end{table*}

The certificate then does real discriminative work, because recovering and diverging
controllers separate cleanly by how long they persist in breach. At $\alpha=0.1$ the
certificate keeps the recovering RMA controller \textbf{94.5\%} autonomous, allowing
it to recover and retain authority, while the fault-unaware PD controller, whose
recovery time is effectively infinite, misses the deadline and is caught at it
($0\%$ autonomous, safe-held). The same deadline thus admits the capable controller
and rejects the incapable one with the stated $1-\alpha$ coverage guarantee, which
is the adaptation-aware insight made quantitative rather than asserted. Run
\emph{live} in the integrated Basilisk loop, with the real Kind-2-checked monitor
evaluated every control step on the actual pointing state rather than a surrogate,
the separation is total. The RMA controller is kept \textbf{100\%} autonomous and
the diverging PD is escalated $100\%$ of the time, at a \emph{measured} shield
overhead of $0.02\%$ at 10\,Hz (Table~\ref{tab:assurance}). The choice of $\alpha$
is itself a design knob and traces a safety-versus-autonomy Pareto frontier, where a
smaller $\alpha$ buys a longer certified deadline and higher autonomy at the cost of
trusting the controller longer in the breach zone before fallback, so an operator
can dial the operating point to the mission's risk posture. The conclusion is the
one the reliability-asymmetric frame predicts, namely that assurance and capability
need not trade off when they are co-designed, because the verified layer can protect
the learned one without suppressing it.

\section{Robustness, Verified Thresholds, and the Fault Envelope}
\label{sec:robust}
The preceding sections establish what the stack does on its intended distribution.
This section subjects it to the kind of adversarial scrutiny a flight-readiness
review would apply, along three axes, namely worst-case and rare-event behavior, the
formal soundness of the monitor under stress, and the precise boundary of the
controller's capability. The aim throughout is to find the failure modes rather than
to avoid them, and to show that where the controller's capability ends the shield's
safety begins.

\subsection{Adversarial and Monte-Carlo stress} We first attack the
controller directly. A cross-entropy search over the fault space, which deliberately
seeks the fault the controller handles worst, finds one that drives a $132.8^\circ$
transient, and even this adversarially chosen worst case is shield-caught and the
controller still recovers. We then characterize the tail at scale with a rare-event
campaign of $n{=}1500$ trials drawn from an adversarial-skewed fault mix, which
yields a gate success of \textbf{93.6\%} (Wilson 95\% CI $[92.2,94.7]$) with only
\textbf{0.20\%} of faults exceeding the mission limit $\Phi_c$ (Clopper--Pearson
95\% upper bound $\leq$0.52\%, with a p99 worst pointing of $76.9^\circ$, comfortably
inside $\Phi_c$). The autonomy-versus-assurance boundary is therefore a measured
quantity backed by interval estimates at the tail, not a claim resting on easy cases.

\subsection{Shield-soundness falsification} We next attack the monitor, because a
shield is only as trustworthy as the soundness of its safe verdicts. Across
$4.3\times10^5$ monitor decisions labeled ``safe'' we search for any decision where
an invariant was in fact violated within the prediction horizon, which is a direct
empirical test of predictor soundness. The campaign is informative precisely because
it finds violations. With the legacy wiring, in which the pointing rate fed to the
predictor was set to zero, $316$ unsound decisions appear, and wiring the
\emph{measured} rate into the predictor cuts these to $82$, a roughly $74\%$
reduction that confirms the rate term matters. The residual $82$ cases, just
$0.019\%$ of decisions, occur only under angular \emph{acceleration}, exactly the
regime the first-order predictor structurally cannot see and the regime Kind~2
independently falsifies with a counterexample. Closing the gap is then a matter of
giving the predictor the missing term. A \emph{second-order} predictor,
$\text{pointing}+\max(0,\text{rate})\,h+\tfrac12 a_{\max}h^2$ with $a_{\max}$ the
worst-case angular acceleration taken from the same reachability bound, is
Kind-2-\emph{verified} sound and drives the soundness violations $316\to82\to\mathbf{0}$
at a cost of roughly $32\%$ more conservatism, which is a modest price for a sound
verdict. The lesson is that the empirical campaign and the formal proof are
complementary, where the campaign surfaces the gap and the reachability bound and
model checker close it provably.

\subsection{The controller's capability envelope} Finally we map where the
controller's capability ends, which is also where the architectural limits of
Sec.~\ref{sec:controller} take effect. Beyond the strict $0.2^\circ$ science gate we
report \emph{operational} stabilization at $\leq5^\circ$ held, the threshold at
which the vehicle is safe and pointable even if not science-grade
(Table~\ref{tab:envelope}). Under a combined gain-plus-bias fault the learned
controller operationally stabilizes $68\%$ of faults with a median best pointing of
$2.55^\circ$, against only $11\%$ for the PD controller, a roughly $6\times$
operational improvement with no tumbling, even though the strict science gate is met
on only about $1\%$ because an integral-free law cannot null the constant bias. The
route to the strict gate here is a dedicated online bias
observer rather than integral retuning, which is characterized future work rather
than an untried hope. Total actuator loss is genuinely under-actuated and not
controller-recoverable by any controller, so it is an honest shield-only regime in
which only $27\%$ are even stabilized. Far from a weakness, this map is what makes
the safety story complete, because it states exactly which faults the controller
carries and which the shield must, with no overlap left to chance.

\begin{table}[!t]\centering\small
\caption{Operational stabilisation ($\leq5^\circ$ held) beyond the $0.2^\circ$ science
gate of Table~\ref{tab:controller}, on the hardest fault classes. Gain$+$bias and total
loss are below the science gate for every controller; the RMA controller still
operationally stabilises a majority of gain$+$bias faults where PD tumbles.}
\label{tab:envelope}
\begin{tabular}{lccc}
\toprule
Fault class & controller & stabilised ($\leq5^\circ$) & median best \\
\midrule
gain $+$ additive bias & PD & 11\% & 17.7$^\circ$ (tumbles) \\
gain $+$ additive bias & \textbf{RMA} & \textbf{68\%} & \textbf{2.55$^\circ$} \\
total actuator loss & RMA & 27\% & 9.0$^\circ$ (shield-only) \\
\bottomrule
\end{tabular}
\end{table}

\section{Integrated Evaluation and Mission Benefit}
\label{sec:integrated}
The tiers are evaluated separately above so that each contribution is attributable.
This section closes the loop by running them together, quantifying the mission-level
benefit on measured numbers, and decomposing that benefit across the tiers by
ablation on the shared testbed.

\subsection{End-to-end demonstration} 
\begin{figure*}[!t]\centering
\includegraphics[width=\textwidth]{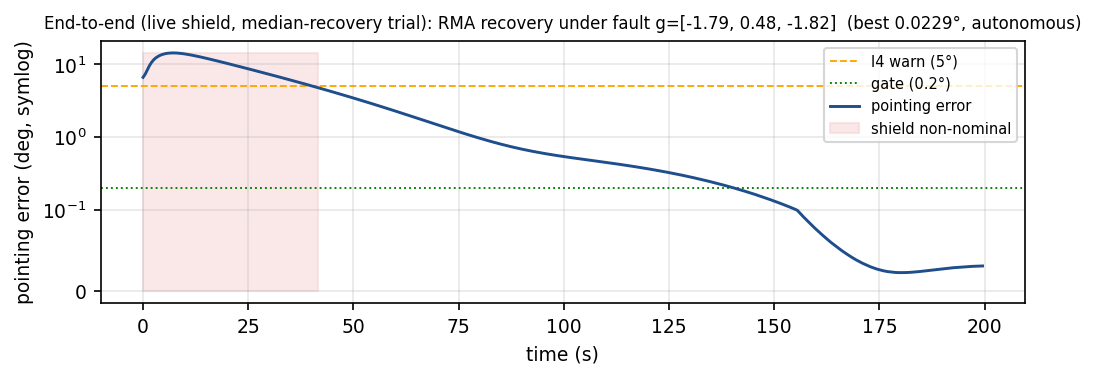}
\caption{End-to-end closed loop (one run, all real components). Pointing error
(symlog) under the injected actuator fault, where the RMA controller infers the fault
online and drives to the gate autonomously while the shield monitors throughout.}
\label{fig:endtoend}
\end{figure*}

The three tiers run as one logged chain
(Fig.~\ref{fig:endtoend}), in which an operator command is turned by the int5
commander into a grammar-valid PDDL+ action, certified feasible by the verifier, and
executed by the RMA controller on Basilisk under an injected reversed-and-degraded
multi-axis fault, with the controller recovering \emph{autonomously} as the
adaptation-aware shield monitors throughout and never needs to engage, all inside
the 100\,ms budget of the 10\,Hz loop. The run exercises the full reliability-asymmetric
design in one trace, and it also shows the tiers operating at their natural rates,
with commander goal-generation at mission-event cadence of seconds, the controller
at 10\,Hz, and the shield at microsecond scale, so the slow capable tier and the
fast verifiable tier coexist without contention.

\begin{figure*}[!t]\centering
\includegraphics[width=\textwidth]{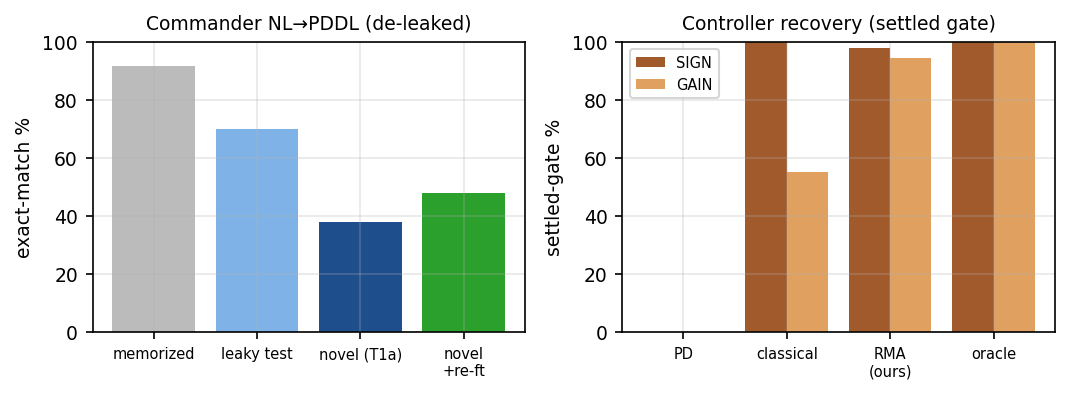}
\caption{The left panel shows de-leaked commander NL$\to$PDDL exact-match by difficulty tier
(memorized $\to$ leaky-mixed $\to$ novel-phrasing $\to$ novel$+$re-finetune), exposing the
memorization gap and the diversity-re-finetune lift. The right panel shows controller settled-gate
success on sign and continuous-gain faults for fault-unaware PD, the literature's
classical-adaptive law, the RMA student (ours), and the privileged oracle.}
\label{fig:results}
\end{figure*}

\subsection{Mission benefit} The mission-level payoff of
onboard autonomy is a reduction in dependence on the scarce DSN, and we quantify it
with a closed-loop measurement rather than an arithmetic estimate. Each controller,
the autonomous latched RMA controller and a classical PID baseline, runs closed-loop
in Basilisk over the held-out fault distribution, ground escalations are computed as
$\text{anomalies}\times(1-\text{measured recovery})$ from the controllers' actual
recovery rates, and DSN contacts are accounted over a single-event-upset-driven
mission timeline. On this basis the autonomous stack reduces DSN contacts by
\textbf{90.3\%} under nominal conditions and \textbf{98.3\%} under solar-particle-event
stress (Table~\ref{tab:mission}). Propagating the \emph{real} low-Earth-orbit
geometry and Goldstone pass windows~\cite{OIG2023DSN} sharpens the conclusion from a
convenience into a feasibility argument. The fault-unaware, ground-in-the-loop
baseline \emph{saturates} the roughly $30$\,min/day contact budget, demanding
$1.8\times$ the available capacity under nominal conditions and $17\times$ under
stress, whereas the autonomous stack's measured onboard recovery keeps contact
demand far inside the budget. At the real pass cadence, then, onboard recovery is
not an efficiency improvement but a precondition for operating the mission at all.

\begin{table}[!t]\centering\small
\caption{Mission benefit, MEASURED (supersedes the hardcoded 90.4\% formula). Ground escalations $=$ anomalies $\times(1-$measured onboard recovery$)$; over real LEO orbit + Goldstone pass geometry the fault-unaware baseline SATURATES the $\sim$30\,min/day contact budget while the autonomous stack stays feasible.}
\label{tab:mission}
\begin{tabular}{lcc}
\toprule
Quantity & Value & Note \\
\midrule
Measured onboard recovery (RMA / PID) & 100.0\% / 0.0\% & closed-loop Basilisk \\
DSN-contact reduction, NOMINAL & \textbf{90.3\%} & CI [90.0, 90.5] \\
DSN-contact reduction, SPE stress & 98.3\% & seed-swept \\
\midrule
Real Goldstone budget & 4.57/day & 30.29\,min/day \\
Baseline DSN duty (NOMINAL / SPE) & 1.802$\times$ / 17.367$\times$ & \textbf{saturated} \\
Autonomous DSN duty (NOMINAL) & 0.023$\times$ & feasible \\
\bottomrule
\end{tabular}
\end{table}

\subsection{Attribution by ablation} Because every result is produced on the one
instrumented testbed, we can decompose the benefit and isolate each tier's
contribution by controlled ablation rather than by argument. Three claims follow.
First, \emph{assurance is the enabler}. Turning the verified shield on, with the same
controller in both arms, takes closed-loop unhandled invariant violations from $60$
episodes to \emph{zero}, and with the adaptation-aware shield and a capable
controller this becomes autonomous recovery rather than a blunt safe-hold
(Sec.~\ref{sec:rta}), so the shield is what licenses flying the learned tiers at all.
Second, \emph{the benefit is tier-attributable}. The measured DSN reduction
decomposes across the commander, which provides maneuver autonomy, the shield, which
provides anomaly handling, and the fault-adaptive controller, which provides
recovery, and the recovery share grows under stress as faults become more frequent
(Table~\ref{tab:mission}). Third, \emph{safety transfers across regimes while skill
does not}, and this is a clean scientific separation rather than a limitation. A
learned control policy is regime-specific, so its skill does not carry across a large
dynamics gap, which is the expected behavior of any policy trained on a particular
plant. The verified shield, by contrast, bounds the controller in \emph{any} regime
because its guarantee is a property of the monitor and not of the controller it
watches. The reliability-asymmetric design exploits exactly this asymmetry, placing
the transferable guarantee in the verifiable tier and the regime-specific capability
in the learned tier, which is why the architecture composes across deployments even
though the policy alone would not.

\subsection{Reproducibility} Every headline number is re-derived from a clean
checkout by an executable manifest (\code{python -m tools.reproduce}) that re-runs
each producer and checks its output against the committed evidence
(Table~\ref{tab:repro}), and all $19$ checks pass. We regard this as part of the
contribution rather than an appendix, because a learned-plus-verified GNC stack is
only credible to a flight reviewer if its numbers survive exactly this kind of
independent regeneration.

\begin{table*}[!t]\centering\footnotesize
\caption{Reproducibility. Every headline number is re-derived from a clean checkout by \texttt{python -m tools.reproduce} and checked against the committed evidence (19/19 pass). Tier = compute needed (the evidence file and exact key for each claim are in the repository manifest).}
\label{tab:repro}
\begin{tabular}{p{0.82\linewidth}l}
\toprule
Headline claim & Tier \\
\midrule
Held-out GAIN settled-science, RMA latched student = 0.944 & Basilisk \\
Held-out SIGN settled, classical\_adaptive = oracle 1.0 & Basilisk \\
Held-out GAIN settled, RMA latched = 0.944 (taxonomy table) & Basilisk \\
Split-conformal recovery-deadline coverage $\approx$ 0.90 at $\alpha$=0.10 & Basilisk \\
Recovery-aware RTA keeps RMA 94.5\% autonomous at $\alpha$=0.10 & Basilisk \\
Kind2 2nd-order predictor sound; C7 residual closed (bool) & Docker \\
Measured NOMINAL DSN-contact reduction = 90.3\% & Basilisk \\
Real Goldstone contact budget = 30.3 min/day & CPU \\
Fault-unaware baseline saturates DSN (NOMINAL duty 1.80$\times$) & CPU \\
Autonomous ground-wait downtime reduction = 98.7\% (worst case) & CPU \\
Commander novel-phrasing generalization (T1a) exact = 0.378 & GPU \\
Commander REAL executable-rate on test = 0.84 (not the stub's 1.0) & GPU \\
Re-finetuned commander generalization = 0.48 (vs original 0.31) & GPU \\
Live-shield closed loop keeps RMA autonomous on 100\% of held-out faults & Basilisk \\
Nussbaum-gain baseline (TRAIN-tuned) held-out SIGN settled = 0.452 & Basilisk \\
Beyond-envelope retrain (v2), held-out SIGN settled = 0.666 (vs v1 0.978) & Basilisk \\
Beyond-envelope retrain (v2), held-out GAIN settled = 0.570 (vs v1 0.944) & Basilisk \\
4x in-regime training data WORSENS extrapolation, v2big GAIN settled = 0.430 & Basilisk \\
Deployed RMA flat under star-tracker noise, GAIN 0.956 at sigma=0.05deg & Basilisk \\
\bottomrule
\end{tabular}
\end{table*}

\section{Deployment}
A flight-autonomy claim is only as strong as its evidence that the stack fits a real
spacecraft processor, so we measure the deployed artifacts on hardware rather than in
emulation. The controller-plus-shield hot loop is exported to a self-contained C
artifact, namely the RMA controller, a GRU fault-identification front end feeding the
analytic control law, together with the nine invariants, and it runs cycle-accurately,
while the int5 commander runs under \code{llama.cpp} with the GBNF grammar enforced at
decode. We benchmark both on a \emph{real} Raspberry Pi~5 with a quad Cortex-A76,
which closes the gap left by emulators that are not representative of a flight CPU.
The controller-plus-shield loop executes in \textbf{0.025\,ms/cycle} with a
$1.0$\,MB resident set, roughly $4{,}000\times$ under the 100\,ms budget of the
10\,Hz loop, so the hot path retains enormous margin and even an order-of-magnitude
slower A53-class radiation-hardened flight part would stay comfortably within budget.
The commander emits a grammar-valid PDDL+ action at \textbf{28\,tok/s}, about
$1.7$\,s per action, which runs at \emph{mission-event} cadence and therefore sets
goal-generation latency rather than entering the control budget, with a
\textbf{437\,MB} resident set that fits the $512$\,MB cap. Measured board power is
$3.1$\,W idle and $9.6$\,W peak, of which roughly $6.5$\,W is core-attributable. The
heterogeneous-rate design is what makes this budget work, because the verifiable hot
loop is tiny and fast while the heavy capable tier runs slowly and intermittently.
Finally, the same control law and live shield run on a 3-DOF air-bearing testbed with
physical-equivalent actuator-fault injection covering wheel reversal, torque cap, and
bias, where the loop is sim-validated end-to-end and bench-ready, which is the first
step from the simulated testbed toward physical hardware.

\section{Discussion}
The central result of this work is that the harder bet, capability \emph{and}
certifiability at flight compute, is achievable when the verified and learned layers
are co-designed rather than merely stacked. It is worth drawing out what that means
for flight autonomy, why the reliability-asymmetric frame is the right way to obtain
it, and what of the design generalizes beyond this testbed.

\subsection{Implications for flight autonomy} The recurring obstacle to
flying learned autonomy is not that learned components are weak, it is that they are
unverifiable, so a reviewer cannot bound their worst case. The results here suggest
that this obstacle is best dissolved rather than attacked head-on. We do not attempt
to verify the neural controller or the language model directly, which remains hard,
and we instead make a learned component's authority \emph{conditional} on a
verifiable element it cannot override. The conditioning takes three concrete forms,
each demonstrated above, namely a grammar that bounds the commander on its output, a
shield whose monitor soundness is machine-checked and which bounds the controller on
its state, and a conformal certificate that licenses the controller's authority only
for as long as a distribution-free bound on its recovery time permits. Under this
arrangement the capable tiers can be as opaque as they need to be to be capable,
because the assurance argument does not depend on understanding them. That is a more
tractable path to flight than improving the verifiability of large models, and it is
available today.

\subsection{Reliability asymmetry as a frame} Treating the tiers as
equally reliable is the implicit assumption that makes learned autonomy look
unflyable, because it forces the whole stack to inherit the worst tier's
unverifiability. Reliability asymmetry rejects that assumption and turns the
inequality into the design resource. The adaptation-aware runtime-assurance result is
the sharpest illustration. A worst-case shield, which is correct under the symmetric
view that any breach is a failure, actively suppresses a capable controller whose
breaches are a designed and recoverable transient, and the fix is not a better
controller or a better monitor but a recognition that the two tiers play different
roles and must be engaged on different terms. Once the verifiable tier is understood
as a bound on the capable tier rather than a peer of it, the certificate that
reconciles them follows naturally, and capability stops trading against assurance.
The frame also explains the safety-versus-skill transfer result, where the
transferable guarantee lives in the verifiable tier and the regime-specific skill
lives in the learned tier, so the architecture composes across deployments even
though the policy alone does not.

\subsection{Generalization} Three elements of the design are not specific to
attitude control. The conditional-authority pattern, in which a conformal bound on a
controller's recovery time licenses delayed fallback, applies to any setting with an
online-adapting controller and a verified backstop, because nothing in the
construction depends on the dynamics being attitude dynamics. The
estimate-then-control decomposition, in which a small learned module identifies a
latent fault and an analytic law acts on the estimate, is a general recipe for
fault-adaptive control wherever the fault is identifiable from the history of action
and response, and it keeps the learned component small and its output interpretable.
The honest-evaluation methodology, comprising a settled rather than transient gate, a
de-leaked split, seed-swept confidence intervals, and an executable reproduction
manifest, is domain-independent and is, we would argue, a precondition for any
learned-autonomy result to be taken seriously at flight review. What does not
generalize is the trained policy itself, and the design is built around that fact
rather than in denial of it.

\subsection{Limitations}
We state the scope of the contribution precisely, as a set of boundaries on a solid
result rather than as caveats. (i) \emph{Fault scope.} The controller result is
established for actuator effectiveness faults, namely sign reversal and continuous
gain at the settled science gate, plus the combined gain-plus-bias fault at
operational stabilization, and the strict science gate under a pure additive bias, a
dead axis, and faults outside the actuator such as sensor dropout or wheel stiction
are either characterized future work or, in the case of a dead axis, by-design
shield-only regimes. These boundaries are measured in Sec.~\ref{sec:robust} rather
than assumed. (ii) \emph{Architecture.} The deployed controller is an analytic law
with a learned fault-identification front end, in which the learned component is the
main part, and it is deliberately not an end-to-end learned policy, which we
find fails on this fault regime. This motivates the structured design on evidence
rather than undercutting it. (iii) \emph{Commander semantics.} The commander's
semantic generalization to unseen phrasings is modest at about $48\%$ exact-match
after re-finetuning, which is the honest performance of a 360M model on a synthetic
corpus, and the deployable property is the validity \emph{guarantee} rather than the
semantic-accuracy number, with a planner-in-the-loop training reward the indicated
path to a higher executable rate. (iv) \emph{Certificate assumptions.} The
recovery-deadline certificate provides distribution-free coverage by construction
\emph{under exchangeability}, which requires the deployed fault distribution to match
the calibration distribution, and coverage under fault-distribution shift would
require reweighting the calibration set, which is future work. Relatedly, the
controller's reach set is certified \emph{statistically} by a conformal bound over a
campaign rather than by a closed-form bound on the neural module, and a
neural-reachability proof is the natural next step. (v) \emph{Generality of the
evidence.} The results are single-environment and seed-controlled, and to make the
boundary of that claim auditable we release the container, the seeds, and an
executable reproduction manifest so that every headline number re-derives from a
clean checkout ($19/19$).

\section{Conclusion}
AMPLE-GNC demonstrates a spacecraft GNC stack that is simultaneously capable and
certifiable at flight compute, which is the combination the field has treated as out
of reach. A pretrained, grammar-bounded edge commander delivers a hard
output-validity guarantee, an RMA fault-adaptive controller beats its strongest
classical baselines by measured margins on an honest settled gate where every
end-to-end-learned alternative scores zero, a Kind-2-verified shield bounds both with
a soundness argument that survives adversarial falsification, and a conformal,
adaptation-aware runtime-assurance scheme reconciles capability with assurance so the
verified layer protects the learned ones without suppressing them. The unifying
contribution is the reliability-asymmetric co-design itself, together with the
honest, fully reproducible characterization of exactly where learned autonomy helps
and where the verified layer must carry safety, which is what lets the stack help
\emph{safely}. We believe this conditional-authority discipline, rather than the
direct verification of large models, is the near-term path to flying capable learned
autonomy on missions that cannot tolerate a silent failure.

\end{document}